\documentclass{article}
\usepackage{ijcai26}

\usepackage{times}
\usepackage{soul}
\usepackage{url}
\usepackage[hidelinks]{hyperref}
\usepackage[utf8]{inputenc}
\usepackage[small]{caption}
\usepackage{graphicx} 
\usepackage{amsmath}
\usepackage{amsthm}
\usepackage{booktabs}
\usepackage{algorithm}
\usepackage{algorithmic}
\usepackage[switch]{lineno}

\usepackage{multirow}
\usepackage{amssymb}
\usepackage{csquotes}
\usepackage{color}
\usepackage{mwe}
\usepackage{cuted}
\usepackage{subcaption}
\usepackage{adjustbox}
\usepackage{xcolor}

\title{Online Operator Design in Evolutionary Optimization for Flexible Job Shop Scheduling via Large Language Models}

\author{
Rongjie Liao$^1$
\and
Junhao Qiu$^{2*}$\And
Zhenguo Yang$^1$\And
Xin Chen$^{2}$\And
Xiaoping Li$^1$\\
\affiliations
$^1$School of Computer Science, Guangdong University of Technology\\
$^2$Department of Computer Science, City University of Hong Kong\\
\emails
liaorongjie1@mails.gdut.edu.cn,
junhaoqiu2-c@my.cityu.edu.hk,
yzg@gdut.edu.cn,
xchen3252-c@my.cityu.edu.hk,
xpli@gdut.edu.cn
}

\begin{document}

\maketitle

\begin{abstract}
Customized static operator design has enabled widespread application of Evolutionary Algorithms (EAs), but their search effectiveness often deteriorates as evolutionary progresses.
Dynamic operator configuration approaches attempt to alleviate this issue, but they typically rely on predefined operator structures and localized parameter control, lacking sustained adaptive optimization throughout evolution.
To overcome these limitations, this work leverages Large Language Models (LLMs) to perceive evolutionary dynamics and enable operator-level meta-evolution. 
The proposed framework, \underline{LLMs} for online operator design in \underline{E}volutionary \underline{O}ptimization, named LLM4EO, comprises three components: knowledge-transfer-based operator design, evolution perception and analysis, and adaptive operator evolution. 
Firstly, operators are initialized by leveraging LLMs to distill and transfer knowledge from well-established operators.
Then, search behaviors and potential limitations of operators are analyzed by integrating fitness performance with evolutionary features, accompanied by suggestions for improvement.
Upon stagnation of population evolution, an LLM-driven meta-operator dynamically optimizes gene selection of operators by prompt-guided improvement strategies.
This approach achieves co-evolution of solutions and operators within a unified optimization framework, introducing a novel paradigm for enhancing the efficiency and adaptability of EAs.
Finally, extensive experiments on multiple benchmarks of flexible job shop scheduling problem demonstrate that LLM4EO accelerates population evolution and outperforms tailored EAs.

\end{abstract}

\begin{figure}[htbp]
  \centering
  \begin{subfigure}[b]{0.4\columnwidth}
    \centering
    \includegraphics[width=\linewidth]{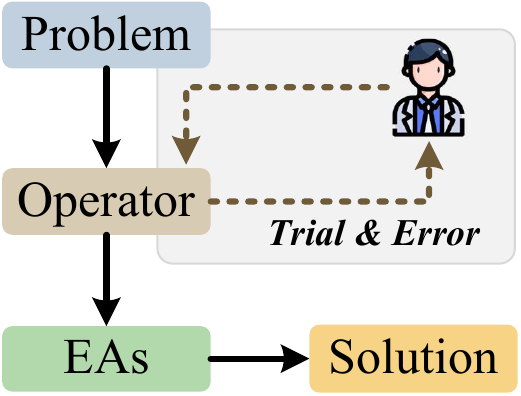}
    \caption{Manual Design}
    \label{fig1 (a)}
  \end{subfigure}\hfill
  \begin{subfigure}[b]{0.4\columnwidth}
    \centering
    \includegraphics[width=\linewidth]{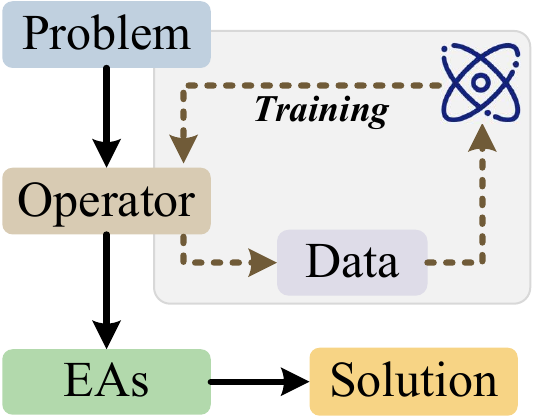}
    \caption{Offline Auto-design}
    \label{fig1 (b)}
  \end{subfigure}%
  \vspace{1em}
  \begin{subfigure}[b]{0.83\columnwidth}
    \centering
    \includegraphics[width=\linewidth]{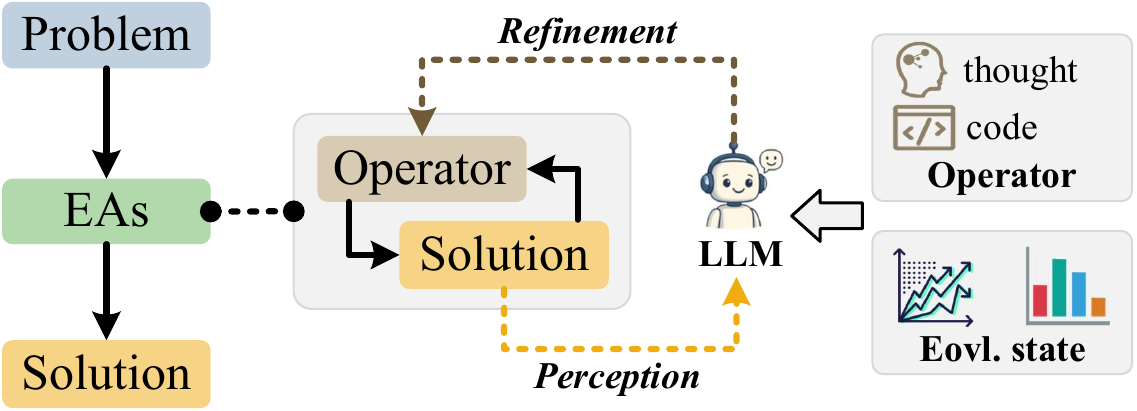}
    \caption{Online Automatic Operator Design}
    \label{fig1 (c)}
  \end{subfigure}
  \caption{Traditional operator design (a) relies on expertise knowledge and trial-and-error; offline automatic design (b) method trains operators on validation data before solving target problems; while our approach (c) online evolves operators by LLMs in the search.}
\end{figure}

\section{Introduction}

The differentiated design of operators has enabled Evolutionary Algorithms (EAs) to be widely applied in fields, such as production scheduling, logistics optimization, and medical resource allocation \cite{r:31}. 
Traditional operator design, relies on expert knowledge to balance exploration and exploitation by controlling the search direction and intensity of neighborhood solutions\cite{r:33}.
However, this search paradigm typically employs fixed operator structures and static parameter configurations, making it difficult to adapt to evolutionary dynamics, such as changes in population structure, neighborhood space and feature distribution.
As a result, the adaptability and generalization of search operators are limited in complex and diverse scenarios \cite{r:61}. This raises an intriguing question: \textit{\textbf{How to utilize evolutionary information to achieve operator evolution during the search process?}}

Dynamically adjusting the search paradigm is crucial for improving the adaptability of EAs and addressing the limitations of statically designed search operators.
Existing dynamic operator configuration approaches mainly include parameter control methods and evolutionary procedure induction algorithms. Although parameter control methods guide the search by adjusting initial and process parameters \cite{r:30}, their effectiveness is fundamentally constrained by the rigid structure of the operator. In addition, evolutionary program induction algorithms \cite{r:20}, such as Genetic Programming and Genetic Evolutionary Programming, search for operators expressed by symbols within an evolutionary framework. However, these algorithms often suffer from inefficient search due to excessive iterations, and the resulting operator quality is highly dependent on validation sets. This limitation arises mainly from two key issues: \textbf{1) Evolutionary information is challenging to be perceived and utilized to guide operator evolution:} Static strategies are incapable of simultaneously meeting the optimization demands of different search stages. 
\textbf{2) Lack of operator-level evolutionary adaptation mechanisms:} Existing approaches primarily rely on local parameter tuning or predefined operator structures, which prevents search operators from being continuously refined based on evolutionary feedback during the optimization process.

To bridge these research gaps, we propose a novel \textbf{large language model for online operator design in evolutionary optimization framework}, termed \textbf{LLM4EO}, which enables operator-level meta-evolution in evolutionary algorithms.  
By leveraging the semantic understanding capabilities of Large Language Models (LLMs), LLM4EO perceives population-level evolutionary dynamics and guides the adaptive refinement of search operators.
In contrast to methods that directly construct solutions \cite{r:28} or generate complete solving algorithms \cite{r:35}, LLM4EO focuses on dynamically analyzing and improving operators throughout the search process.
When population evolution stagnates, an LLM-driven meta-operator is activated to adaptively modify the gene selection strategies and behavioral patterns of underlying operators, thereby accelerating convergence and enhancing the adaptability of EAs.

We further design an LLM-driven meta-operator and a collaborative evolution architecture that couples the evolution of solutions with the adaptive refinement of operators.
The framework consists of three tightly integrated components.
\textbf{1) Knowledge-transfer-based operator design}, where the LLM generates a high-quality initial operator population by leveraging prior knowledge of problem structures and classical heuristics.
\textbf{2) Evolution perception and analysis}, where fitness indicators of both solutions and operators are evaluated to capture critical evolutionary information, including population distribution, convergence trends, and performance variations across search stages. Based on these signals, the LLM perceives the current evolutionary state and diagnoses potential deficiencies in operator behaviors.
\textbf{3) Adaptive operator evolution}, where the LLM-driven meta-operator dynamically updates the gene selection strategies and behavioral patterns of underlying search operators through prompt-guided optimization.
These components are integrated via a structured prompt framework, forming a closed-loop process of perception, analysis, and refinement that continuously improves operator effectiveness during the evolutionary search.
Furthermore, the effectiveness of LLM4EO is validated on multiple benchmark datasets of the flexible job shop scheduling problem.
Experimental results demonstrate that LLM4EO consistently outperforms mainstream methods in both convergence speed and solution quality, particularly when the search process encounters stagnation or local optima.
Overall, this work illustrates how LLMs can be systematically leveraged to enable operator-level adaptive refinement, offering a new perspective on enhancing the performance and adaptability of evolutionary algorithms across diverse optimization scenarios.

\section{Related Work}
\subsection{Operator Control and Design.}
Operators are central to balancing exploration and exploitation in EAs, directly determining convergence and optimization performance.
Heuristic operators traditionally designed to rely on expert knowledge and iterative trials achieve excellent performance in specific problems, but have limited generalization capabilities.
Enhancing the adaptability of the operator to the dynamic features of the search process has become the focus of operator control and design.
Parameter control methods as one of the solutions can select appropriate parameter combinations for different search stages \cite{r:19}, such as crossover/mutation rates \cite{r:15,r:16}, step sizes \cite{r:17}, strategy selection \cite{r:14}, and individual selection \cite{r:13}, but still constrained by human-defined rules. 
Another class of solutions is the use of evolutionary program induction algorithms to construct dynamically adaptable operators. Neighborhood strategy optimization is implemented by automatically evolving rules in a finite terminal symbol space, e.g., Genetic Programming (GP) and Gene Expression Programming (GEP)~\cite{r:22,r:23}. 

Recent integrations of LLMs into EAs primarily follow two paradigms: as direct search operators \cite{r:28,r:29,r:58} and as offline strategy designers \cite{r:53,r:64}. In the first paradigm, leveraging LLMs for high-frequency solution generation often incurs prohibitive computational latency and execution overhead \cite{r:37}. Conversely, the second paradigm relies on a protracted pre-training phase, rendering algorithmic efficacy contingent upon validation data quality \cite{r:63}. Consequently, integrating online operator automated design that perceives and responds to dynamic search features is critical for flexible neighborhood adjustments. This paradigm eliminates offline training dependencies and facilitates state-aware responses to search dynamics, thereby balancing execution efficiency with adaptability.



\begin{figure*}[h]
\centering
\includegraphics[width=0.95\textwidth]{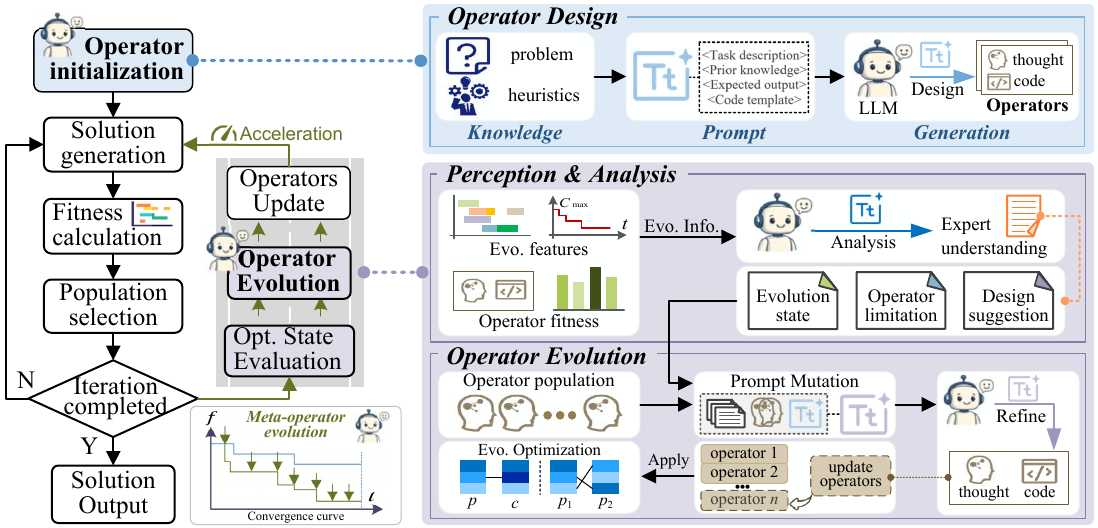}
\caption{The framework of LLM4EO.}
\label{fig2}
\end{figure*}

\subsection{Automatic Heuristic Design.}
Automatic heuristic design (AHD) selects, tunes, or constructs effective heuristics, addressing application bottlenecks of traditional manual design that relies on labor-intensive and expert experience \cite{r:27}.
GP has been applied to AHD and generates heuristics by crossover and mutation of a set of allowed symbols, requiring precise fitness landscapes and expensive search costs \cite{r:21}.

Due to the powerful comprehension and generation capabilities of LLMs, prior works have used them to design heuristics in an evolutionary search framework \cite{r:26,r:27,r:39,r:58,r:59,r:60}. Specifically, EOH introduces thoughts to achieve co-evolution with the codes and achieves outstanding performance on several classical problems \cite{r:27,r:37}. Furthermore, \cite{r:58} introduces fitness landscape analysis and black-box prompting via the added LLM agent for reliable heuristic evaluations.
The integration of LLMs and evolutionary computation introduces a novel algorithmic design paradigm. Their capabilities in summarizing and refining semantic knowledge provide new insights for the continuous online design of operators within complex optimization spaces.


\section{Problem}
The flexible job shop scheduling problem (FJSP) involves scheduling $I$ jobs, where each job $i \in \{1,...,I\}$ consists of  $J_i$ operations that must be processed in a specific sequence. Each operation $O_{i,j} (j = 1,..,J_i)$ can be performed on any machine $m$ in a set of eligible machines $M_{i,j}$, with processing time $T_{i,j,m}>0$. The start and finish time of operation $O_{i,j}$ are $S_{{i,j}} \geq 0$ and $F_{{i,j}} = S_{{i,j}} + T_{i,j,m} \times x_{{i,j,m}}$, respectively. The binary variable $x_{{i,j,m}}=1$ if operation $O_{{i,j}}$ is processed in machine $m$, and 0 otherwise. The optimization objective is to minimize the makespan $C_{max}$.

\section{LLM4EO}
The proposed framework, LLMs for online operator design in evolutionary optimization (LLM4EO), leverages the LLM to generate and refine operators in the search of EAs, thereby enhancing the algorithmic performance, as shown in Figure \ref{fig2}. There are three core components: 1) knowledge-transfer-based operator design, 2) evolution perception and analysis, and 3) adaptive operator evolution. First, the LLM is used to design gene selection strategies of operators based on prior knowledge, then construct a high-quality initial operator population to generate new solutions. To prevent premature convergence, a threshold is set to trigger the adaptive operator evolution mechanism. The LLM is employed to perceive the evolutionary state of solution population, as well as to analyze the search preferences and limitations of operator population. The resulting suggestions guide the LLM to create new promising operators, thus accelerating the search process. LLM4EO facilitates the co-evolution of solutions and operators, providing an intelligent optimization approach.

\subsection{Solution Representation and Initialization}
In LLM4EO, Genetic Algorithm (GA) is adopted to solve the FJSP, where solutions are encoded into two parts: operation sequence vector (OSV) and machine assignment vector (MAV). Both vectors are arrays of integers, the length of which is the total number of operations.

The initial solution population is generated by two machine assignment rules and three operation dispatching rules \cite{r:8}. The fitness of each solution is calculated as $f_{soln}=1/C_{max}$, ensuring the shorter makespan corresponds to the higher fitness. To speed up convergence, the tournament selection method is used to select individuals to generate solutions. After each iteration, the best individual is retained and directly enters the next generation.

\subsection{Meta-operator}
To reduce dependence on manual design and enhance evolutionary adaptability, we propose a meta-operator that employs an LLM-driven gene selection strategy to dynamically explore high-fitness regions of the search space while generating better solutions through neighborhood moves. 

\paragraph{Neighborhood Moves.} 
In this work, four typical neighborhood moves are adopted as structural transformation operations on selected genes to generate new solutions. 

1) OSV crossover: 
The Precedence Preserving Order-based crossover (POX) \cite{r:9}. First, a job is selected from the first parent, and all its operations are copied to the first offspring. Then, the remaining operations are filled in according to the operation sequence of the second parent.

2) MAV crossover: 
Select some operations and exchange their machines between the two parents.

3) OSV mutation: 
The Precedence Preserving Shift mutation (PPS) \cite{r:9}. The sequence of operations is changed by moving one of them to a different position, while satisfying the process path constraints.

4) MAV mutation: 
Select an alternative machine to replace the original machine assigned to an operation.

5) Critical operation swapping:
The critical path is the longest path in the scheduling scheme. Two critical operations are selected to swap. If a better solution is found, the local search process continues along the new critical path.

\paragraph{Gene Selection.}
Gene selection strategies define the search scope and direction by identifying key decision variables to adjust, thereby guiding the search toward more promising regions. A gene is selected if its probability $\gamma$ is greater than the random number $u$ in [0, 1]. Considering the characteristics of FJSP, we design a hierarchical gene selection method based on jobs and operations, which is defined as:
\begin{equation}
\delta=
\begin{cases}
\{O_{i,j} \mid \gamma_i > u_i, \forall i,j \leq J_i \} \\
\{O_{{i,j}} \mid \gamma_{{i,j}} > u_{{i,j}}, \forall i,j \} \\
\end{cases}
\label{eq1}
\end{equation}
If operation $O_{i,j}$ is selected, it is added to the gene set $\delta$. Moreover, if job $i$ is selected, all of its operations are included. In crossover, offsprings $s'_x$ and $s'_y$ are generated by exchanging selected genes $\delta_x$ and $\delta_y$ of individuals $s_x$ and $s_y$:
\begin{equation}
s'_x,s'_y = crossover(s_x,s_y,\delta_x,\delta_y)
\label{eq3}
\end{equation}
In mutation, the genes $\delta$ of individual $s$ are modified, resulting in offspring $s'$:
\begin{equation}
s' = mutation(s,\delta)
\label{eq4}
\end{equation}



The gene selection controls the size of the search neighborhood to be explored, which directly impacts solution quality and algorithm performance. However, random or single-heuristic approaches often lead to poor exploration and early convergence. To address this issue, we use the LLM to construct gene selection heuristics, enabling the algorithm to adaptively explore potential neighborhoods. Gene selection is formulated as a probability function $g(t,P_{op},\mathcal{N})$, where $t$ denotes the iteration, $P_{op}$ is the operator population, and $\mathcal{N}$ represents genetic features. This function maps the hidden correlations among genes to their selection probabilities. Genetic features $\mathcal{N}$ capture the core properties of the FJSP and guide the inference of the model. For job $i$, there are three genetic features: process span $C_i$, minimal process span $C'_{i}$ and number of operations $J_{i}$. For operation $O_{i,j}$, there are four genetic features: start time $S_{{i,j}}$, earliest start time $F_{{i,j-1}}$, processing time $T_{i,j}$ and number of optional machines $| M_{{i,j}} |$. Notably, process span $C_i$ is the time interval between starting the first operation and completing the last operation of job $i$:
\begin{equation}
C_i=F_{{i,J_i}} - S_{{i,1}}, \forall i 
\label{eq5}
\end{equation}
The minimal process span $C'_i$ represents the sum of the shortest processing times for all operations of job $i$:
\begin{equation}
C'_i=\sum\limits_{j \leq J_i}\min\limits_{m \in M_{{i,j}}}T_{{i,j,m}}, \forall i
\label{eq6}
\end{equation}
The processing time $T_{i,j}$ refers to the duration of the operation $O_{i,j}$ on the machine $m$:
\begin{equation}
T_{i,j}=\sum\limits_{m \in M_{{i,j}}}x_{{i,j,m}} T_{{i,j,m}}, \forall i,j
\label{eq7}
\end{equation}

\paragraph{Fitness Evaluation.} 
To guide the evolution toward high-quality solutions, the fitness of operator population is evaluated, and superior operators are preferentially selected using the roulette wheel method. The optimization success rate of an operator is defined as its fitness $f_{op}=n_s/n_v$, where $n_s$ is the number of better solutions it generates and $n_v$ represents how many times it is selected. A higher $f_{op}$ reflects a stronger performance of the operator.

\subsection{Collaborative Evolution Architecture}
Transcending the limitations of static search strategies in traditional EAs, we propose a collaborative evolutionary architecture for operators and solutions that leverages LLMs for experience transfer, perception, analysis, and generation. 

\paragraph{Knowledge-transfer-based Operator Design.}
A structured initial prompt $\mathcal{R}$ is designed to activate prior knowledge of LLMs, enabling the generation of the high-quality initial operator population $P_{op}$, which is denoted as:

\begin{equation}
P_{op} = \{o_n \mid o_n = DesignByLLM(\mathcal{R}),n = 1,2,...,p_{op}\}
\label{eq8}
\end{equation}
where $p_{op}$ is the operator population size and $o_n = DesignByLLM(\mathcal{R})$ represents operator $o_n$ generated by the LLM under the prompt $\mathcal{R}$ that includes a task description, expected output, prior knowledge and code template.

\begin{itemize}
\item Task description: 
First, the target problem and requirements are described, such as \enquote{For the flexible job shop scheduling problem aiming to minimize makespan, \textit{design an operator to calculate the adjustment priority for jobs and operations}.} 
\end{itemize}

\begin{itemize}
\item Expected output: 
The output incorporates the thought $\mathcal{I}$ and the function $\mathcal{F}$ of the operator $o$. To avoid long and noisy responses, the LLM must summarize the thought $\mathcal{I}$ in one sentence within \{\} and give a structured Python function $\mathcal{F}$ using the given code template: 
\end{itemize}

\begin{itemize}
\item Prior knowledge: To improve the quality of initial operators, LLM can refer to classical heuristics $h_c$, such as the Shortest Processing Time (SPT) rule. Moreover, brief descriptions of the task $\mathcal{T}$ and the genetic features $\mathcal{N}$ guide the LLM to get a novel thought $\mathcal{I}$:
\end{itemize}

\begin{equation}
\mathcal{I}=LLM(\mathcal{T},\mathcal{N}, h_c \mid \mathcal{R})
\label{eq10}
\end{equation}

\begin{itemize}
\item Code template: 
The name, input and output of code block are defined, with detailed explanations of meaning and type of each parameter. Given thought $\mathcal{I}$ and code template $\mathcal{D}$, function $\mathcal{F}$ comprises two components: job priority calculation $\mathcal{F}_{x}$ and operation priority calculation $\mathcal{F}_{y}$. Given genetic features $\mathcal{N}$, $\mathcal{F}$ outputs probabilities that are normalized by the cumulative normal distribution function to ensure values in the interval [0, 1]:
\end{itemize}

\begin{equation}
\mathcal{F} = (\mathcal{F}_{x},\mathcal{F}_{y}) =LLM(\mathcal{I}, \mathcal{N},\mathcal{D} \mid \mathcal{R})
\label{eq11}
\end{equation}

\paragraph{Evolution Perception and Analysis.}
Evolutionary changes in distribution of solution population degrade the effectiveness of search strategies, necessitating the identification of when, why, and how to improve operators. To address this, we design the following steps:

1) Convergence evaluation: 
The convergence state is defined as the number of consecutive iterations $\Delta t$ without improvement in the global best makespan, and then introduce a threshold $\theta=1/(\epsilon\times\Delta t)$ to adaptively trigger operator evolution, where $\epsilon$ is a coefficient controlling the frequency of evolution. Based on extensive experimental experience, $\epsilon=0.05$ is selected to balance the frequency of LLM invocation and the overall algorithmic efficiency. In each iteration, if a random number $u$ in [0,1] is greater than the threshold $\theta$, the operator evolution mechanism will be triggered: 

\begin{equation}
o'=RefineByLLM(P_{op}, P_{soln}, \mathcal{R}' \mid u > \theta)
\label{eq12}
\end{equation}
Here, LLMs refine and create a novel operator $o'$ according to information about operation population $P_{op}$, changes in solution population $P_{soln}$, and improved prompt $\mathcal{R}'$.

2) Perception and analysis: 
The pseudo-code for perception and analysis is shown in Algorithm \ref{algorithm1}. To guild the LLM to perceive the evolutionary state of the current search stage, we record changes in solution population $P_{soln}$, such as minimum fitness, average fitness, and their respective change rates since the last evolution of operators. In addition, information about operator population $P_{op}$ is provided, including the fitness and thought of each operator. Then, the task description $\mathcal{T'}$ is given as follows: \enquote{\textit{Characterize the evolutionary state of solution population, describe the limitations of each operator, and provide a suggestion for designing a new operator}}.

\begin{algorithm}[h]
\caption{Perception and analysis}
\label{algorithm1}
\textbf{Input}: solution population $P_{soln}$, operator population $P_{op}$, task description $\mathcal{T'}$   \\
\textbf{Output}: response result $T_{as}$
\begin{algorithmic}[1]
\STATE Create text $T_{soln}$ of changes in $P_{soln}$
\STATE Create text $T_{op}$ of information about $P_{op}$
\STATE $T_{as} = AnalysisByLLM(T_{soln},T_{op},\mathcal{T'})$
\STATE \textbf{return} $T_{as}$
\end{algorithmic}
\end{algorithm}

\paragraph{Adaptive Operator Evolution.}
The pseudo-code for the operator evolution is shown in Algorithm \ref{algorithm2}. A new prompt $\mathcal{R'}$ retains the structure of initial prompt $\mathcal{R}$ and incorporates the response result $T_{as}$ from the perception and analysis. The task description of $\mathcal{R'}$ explicitly emphasizes \enquote{\textit{develop a completely new algorithm distinct from the previous ones.}} To mitigate response inertia caused by a fixed prompt, the LLM further fine-tunes the task description to increase output diversity. Using mutation-based prompting, a potential operator is generated and replaces the worst-performing one, forming a new operator population. With LLM assistance, the adaptive evolution of operators guides EAs to continuously search towards potential regions in the solution space.

\begin{algorithm}[h]
\caption{Operator evolution}
\label{algorithm2}
\textbf{Input}: result $T_{as}$ of perception and analysis, prompt $\mathcal{R}$   \\
\textbf{Output}: new operator $o'$
\begin{algorithmic}[1]
\STATE $\mathcal{R} \leftarrow$ Task description of $\mathcal{R}$ is fine-tuned by LLM
\STATE Improved prompt $\mathcal{R}' = T_{as} \oplus \mathcal{R}$ \\
\STATE Generate a new valid operator $o'$ based on $\mathcal{R}'$
\STATE \textbf{return} $o'$
\end{algorithmic}
\end{algorithm}

\begin{figure*}[h]
\centering
\includegraphics[width=0.95\textwidth]{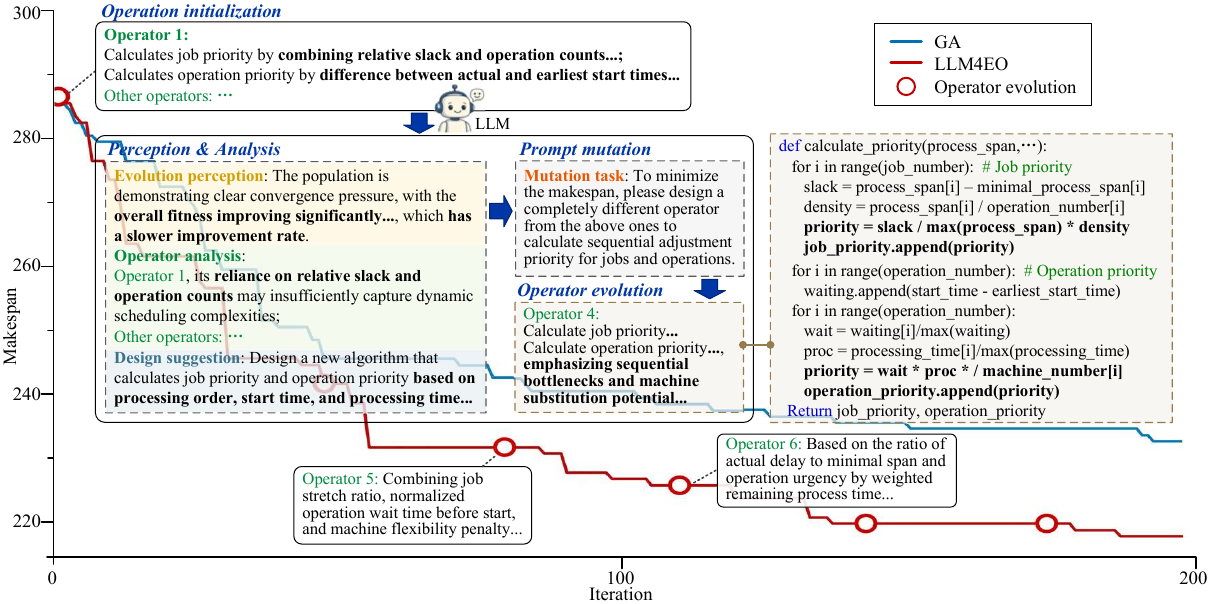}
\caption{Evolution of LLM4EO for FJSP. Key information on operator initialization, perception and analysis, and adaptive operator evolution is presented, including core thoughts and code snippets of operators produced in some generations during evolution.}
\label{Example}
\end{figure*}

\section{Experiments}
\subsection{Experimental Details}
\paragraph{Experimental Design.} 
In this section, extensive computational experiments are conducted to evaluate our proposed LLM4EO. First, the setting of experimental environment and algorithm parameters is described. Then, an ablation study is designed to analyze the impact of each improvement on the algorithm. Subsequently, LLM4EO is compared with the mainstream algorithms for automatic program generation to demonstrate the potential of LLMs in designing operators. Finally, the superior performance of LLM4EO is verified by comparing it with other algorithms.

\paragraph{Parameter Settings.}
The parameters are set as follows: operator population size = 3, solution population size = 100, and maximum iterations = 500. The machine allocation and operation scheduling rules for solution initialization are consistent with \cite{r:8}. The probabilities of crossover and mutation are both 0.9. All experiments are implemented with Python 3.9 and executed on a computer with an Intel core i5-12400 @ 2.50 GHz and 32 GB of RAM.

\begin{table}[h]
 \caption{Comparison results of different LLMs.}
\centering
\setlength{\tabcolsep}{1mm}
\fontsize{9pt}{9pt}\selectfont
\begin{tabular}{cccc}
\toprule
LLM & $RPD$ & Cost & Ratio \\
 \midrule
GPT-4.1-mini	&	36.97	&	\textbf{0.0564\$}	&	\textbf{0.0015} \\
GPT-4o	&	36.57	&	0.1770\$	&	0.0048 \\
DeepSeek-Chat	&	37.58	&	0.2847\$	&	0.0076 \\
Qwen-Max	&	37.78	&	0.3721\$	&	0.0098 \\
Claude-4-Sonnet	&	41.62	&	0.5914\$	&	0.0142 \\
Gemini 2.5 Pro	&	\textbf{36.16}	&	2.7024\$	&	0.0747 \\
\bottomrule
\end{tabular}
\label{table2}
\end{table}

\paragraph{LLM Selection.}
To ensure operator performance, the quality and cost of the algorithm are evaluated in different LLMs, as shown in Table \ref{table2}. The tested models include GPT-4.1-mini, GPT-4o, DeepSeek-Chat, Qwen-Max, Claude-4-Sonnet and Gemini 2.5 Pro. LLM4EO is repeated three times on the MK10 instance with the largest size of the benchmark from \cite{r:2}. Model performance is compared based on three metrics: average relative percent deviation, running cost and quality-price ratio. The relative percentage deviation $(RPD)$ is used to quantify the deviation of each solution from the known optimal solution:
\begin{equation}
RPD=\frac{C_{max}-LB}{LB}\times 100 \%
\label{eq8}
\end{equation}
Here, $LB$ is the lower bound and the lower $RPD$ reflects better algorithm performance. GPT-4.1-mini offers the best cost-effectiveness. Despite its small model size, it achieves relatively satisfactory results, primarily due to the inherent robustness of EAs, which maintain strong performance under limited resources. In contrast, Gemini-2.5-Pro achieves the minimum average $RPD$, but incurs significant costs. Taking into account both algorithmic effectiveness and economic viability, GPT-4.1-mini is considered the most suitable option.

\paragraph{Visualization of Operator Evolution.}
Figure \ref{Example} visualizes the evolution of operators in a running time of the MK10 instance, including the perception and analysis of the first operator evolution, as well as the key thoughts and corresponding codes of operators in different generations. It can be observed that LLM4EO outperforms GA in convergence efficiency, owing to the timely updates of gene selection strategies driven by meta-operators. This enables the algorithm to escape local optima and explore more promising areas. The significant differences in thoughts among various operators and their intricate heuristics reveal the strong comprehension and creativity of LLMs.

\subsection{Performance Comparisons and Analysis}
\paragraph{Ablation Study.}
To evaluate the effectiveness of each improvement, we compare algorithm variants with different components. First, the LLM autonomously designs an initial operator for GA, resulting in a variant algorithm named LLM4OD. Then, the single operator is expanded to an operator population, forming LLM4OPD. Finally, the operator population is further evolved by LLM in the search, yielding LLM4EO. Each instance is solved 10 times to obtain the best makespan ($BM$) and average makespan ($AM$).

Table \ref{table3} shows the $RPD_{BM}$ and $RPD_{AM}$ of instances from \cite{r:2} and \cite{r:62}. The algorithms are ranked from best to worst: LLM4EO, LLM4OPD, and LLM4OD. Notably, LLM4OPD outperforms LLM4OD, as LLM4OD can only select the unique operator per iteration, limiting search scope and reducing population diversity. To balance exploration and exploitation, LLM4EO maintains a diverse operator pool and adaptively generates effective operators tailored to the current search.

\begin{table}[h]
\caption{Comparison results of variant algorithms.}
\centering
\setlength{\tabcolsep}{1mm}
\fontsize{9pt}{9pt}\selectfont
\begin{tabular}{ccccccc}
\toprule
\multirow{2}{*}{Algorithm} & \multicolumn{2}{c}{Brandimarte} & \multicolumn{2}{c}{Hurink-vdata} \\
\cmidrule(lr){2-3} \cmidrule(lr){4-5}
 & $RPD_{BM}$ & $RPD_{AM}$ & $RPD_{BM}$ & $RPD_{AM}$ \\
 \midrule
LLM4OD & 19.11\% & 22.53\% & 3.32\% & 5.05\% \\
LLM4OPD & 18.93\% & 22.33\% & 3.31\% & 4.54\% \\
LLM4EO & \textbf{18.54\%} & \textbf{21.37\%} & \textbf{3.03\%} & \textbf{4.35\%} \\
\bottomrule
\end{tabular}
\label{table3}
\end{table}

\begin{figure*}[htbp]
  \centering
  \begin{subfigure}[b]{0.33\linewidth}
    \centering
    \includegraphics[width=\linewidth]{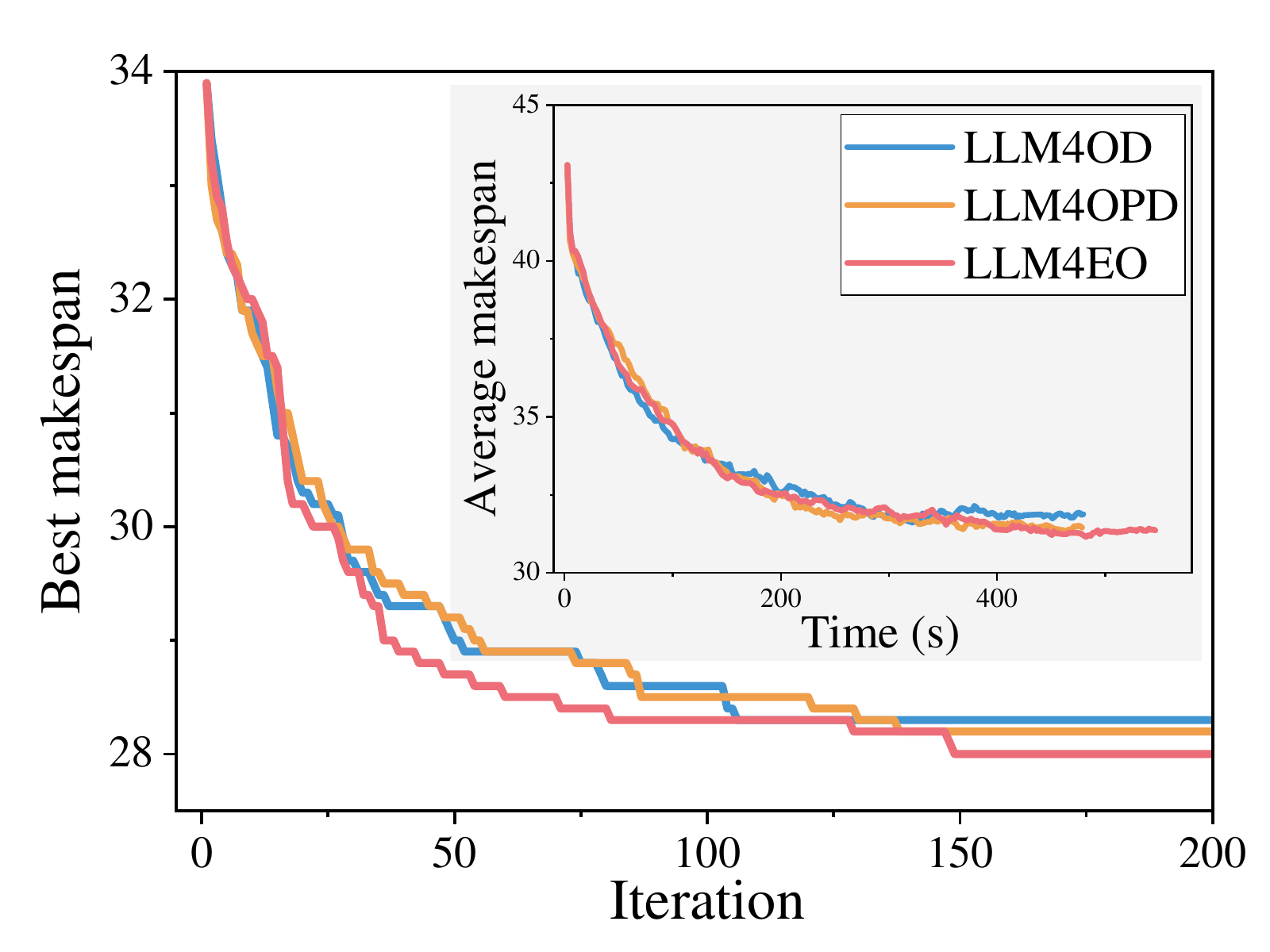}
    \caption{MK02}
  \end{subfigure}\hfill
  \begin{subfigure}[b]{0.33\linewidth}
    \centering
    \includegraphics[width=\linewidth]{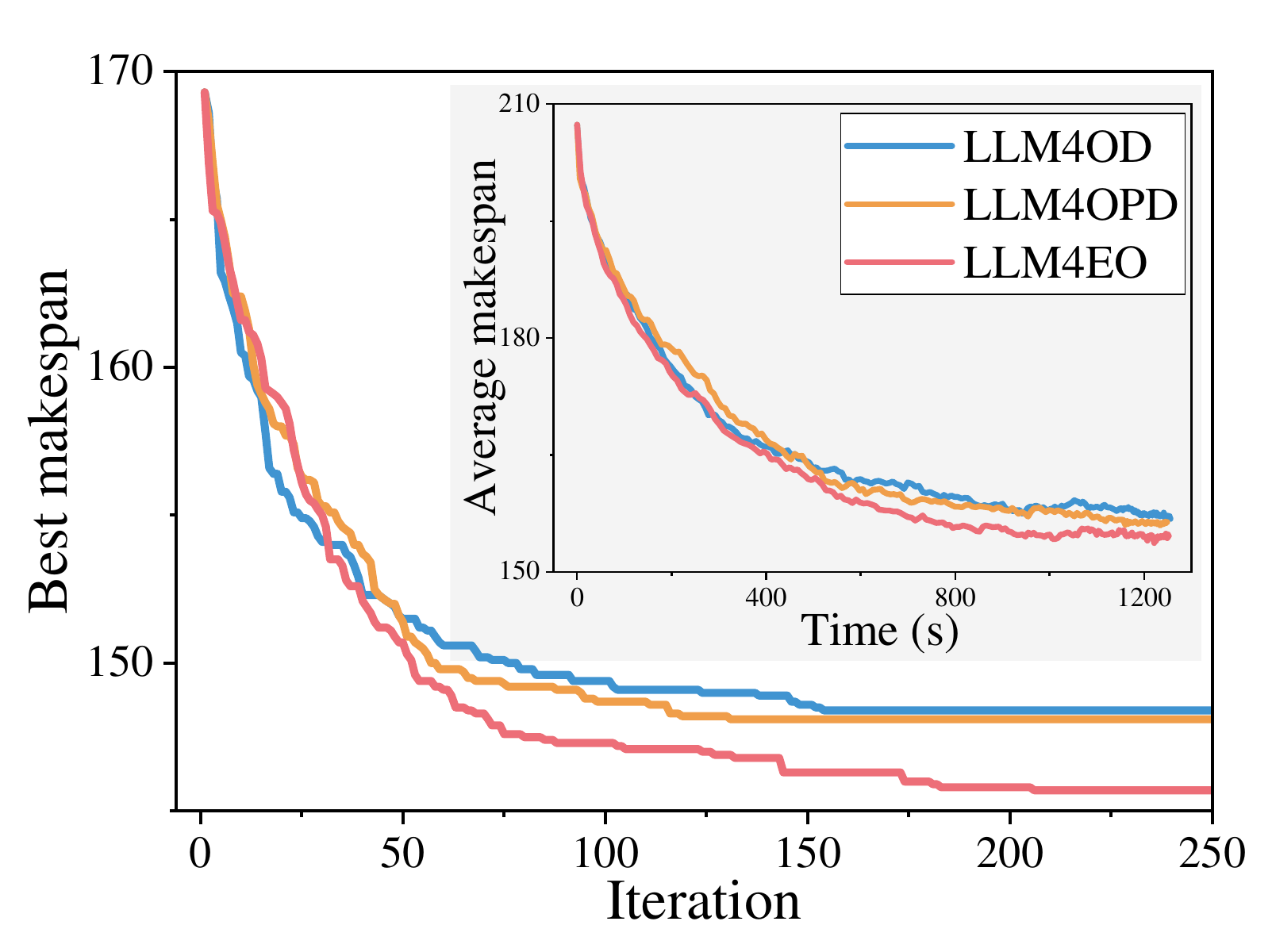}
    \caption{MK07}
  \end{subfigure}
    \begin{subfigure}[b]{0.33\linewidth}
    \centering
    \includegraphics[width=\linewidth]{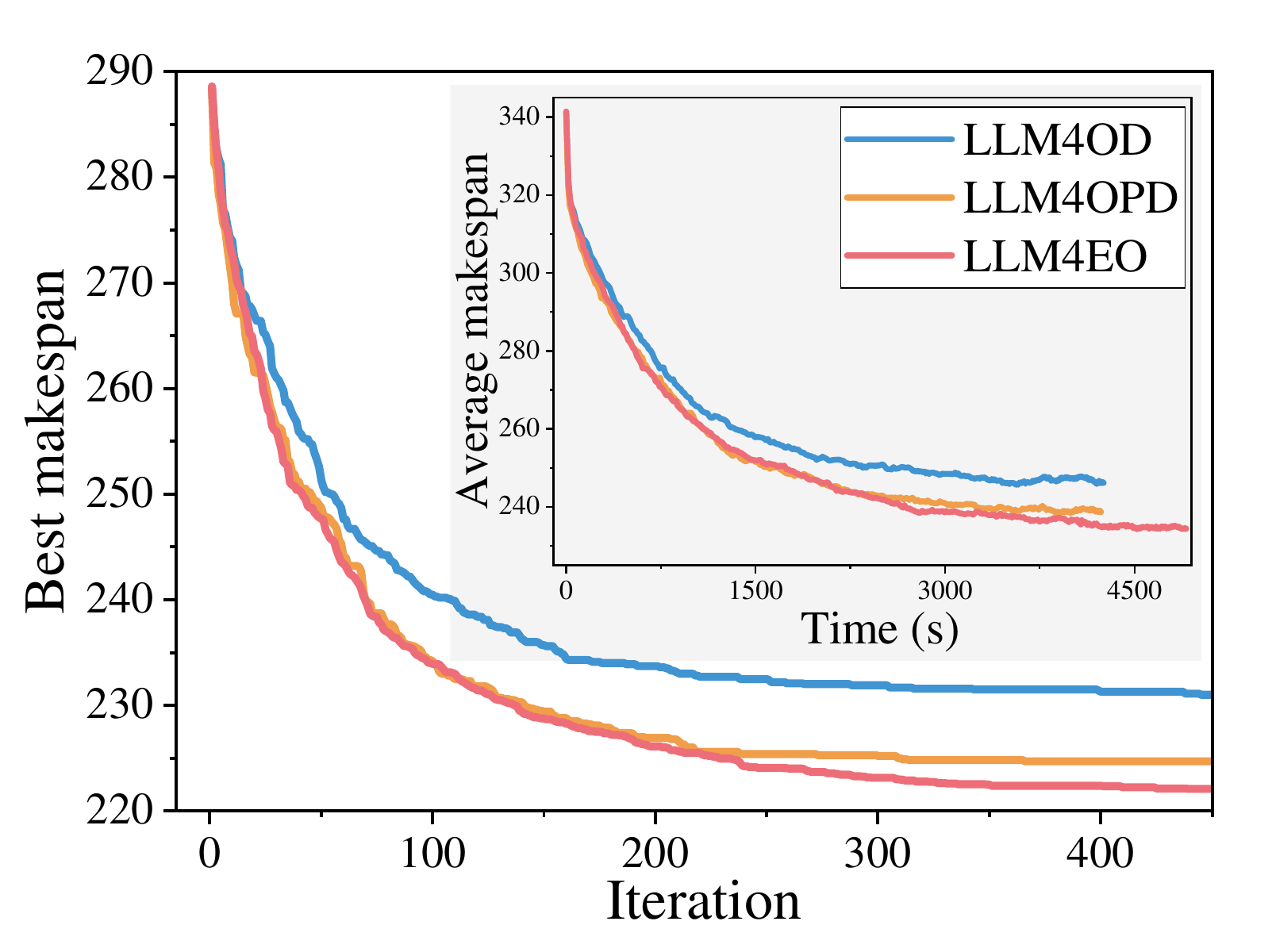}
    \caption{MK10}
  \end{subfigure}
  \caption{Mean convergence curves over the 10 runs for variant algorithms in MK02, MK07, and MK10 of Brandimarte benchmark. The best makespan per iteration and the average makespan over time are shown in each instance.}
  \label{fig3}
\end{figure*}

Figure \ref{fig3} illustrates the convergence curves of MK02, MK07, and MK10, depicting the variation of the best makespan with iterations and the evolution of the average makespan over time. At the same number of iterations, LLM4EO converges faster than other algorithms and finally achieves better solutions. In the time dimension, despite the computational overhead of online operator evolution, LLM4EO demonstrates significant advantages in both the optimization speed and the average objective of the solution population. For larger-scale problems, the proportion of time consumed by operator evolution is significantly reduced. The results show that the collaborative evolution of solutions and operators effectively improves the optimization performance.

\begin{figure}[h]
\centering
\includegraphics[width=0.8\columnwidth]{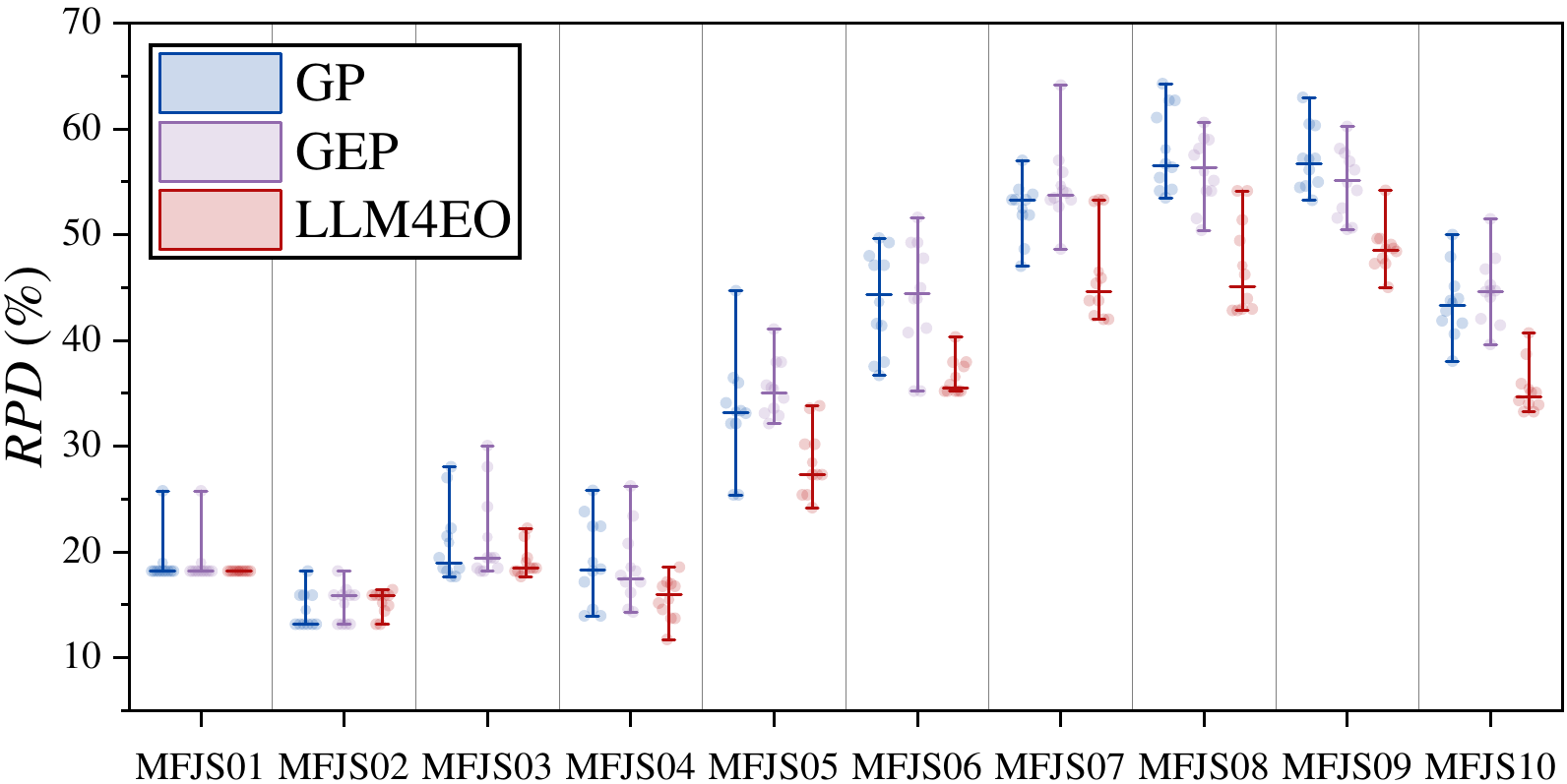}
\caption{Partial box plots of GP, GEP and LLM4EO on Fattahi.}
\label{fig4}
\end{figure}

\paragraph{Comparison of Automatic Operator Design Methods.}
Genetic Programming (GP) and Gene Expression Programming (GEP) are employed to create operators and compared with LLM4EO to assess the potential of LLM in operator design. Since the LLM generates initial operators according to the SPT heuristic, it is also adopted as the initial operator for GP and GEP. To ensure fairness, each method uses the legitimate result produced by a single iteration as the new operator. Each instance from \cite{r:10} is solved 10 times. Since these algorithms find solutions equal to $LB$ of SFJS instances with relatively simple nature, only the box plots of $RPD$ for MFJS instances are shown in Figure \ref{fig4}. 

Figure \ref{fig4} shows that LLM4EO outperforms both GP and GEP in terms of $BM$ and $AM$, demonstrating its superior generation capability. It is closely related to the inherent characteristics of each method. GP and GEP lack semantic understanding, resulting in lower-quality operators without sufficient iterations. In contrast, LLMs have robust representation and understanding, overcoming the limitations of traditional methods in text comprehension and program generation, and exhibiting promising prospects for operator design.

\paragraph{Comparison with Other Algorithms.}
To further explore the performance and generalization of LLM4EO, it is compared with various optimization algorithms such as SLABC \cite{r:44}, SLGA \cite{r:40}, HGIN-RS \cite{r:54}, GRU-DRL \cite{r:55}, HTS/SA \cite{r:10}, AIA \cite{r:50}, IGAR \cite{r:51}, DDEA-PMI \cite{r:38} and the basic genetic algorithm (GA) of LLM4EO. Table \ref{table7} shows the average $BM$ and $RPD_{BM}$ across 10 runs on the datasets. It can be seen that LLM4EO outperforms the compared optimization algorithms. In the same framework, LLM4EO outperforms GA solely through meta-operators. Specifically on Brandimarte dataset, $RPD_{BM}$ decreases from 19.13\% to 18.54\%, resulting in a 3.08\% improvement, demonstrating its effectiveness.

\begin{table}[h]
\caption{Comparison results of other algorithms and LLM4EO.}
\centering
\setlength{\tabcolsep}{1mm}
\fontsize{9pt}{9pt}\selectfont
\begin{tabular}{ccccc}
\toprule
\multirow{2}{*}{Algorithm} & \multicolumn{2}{c}{Brandimarte} & \multicolumn{2}{c}{Fattahi} \\
\cmidrule(lr){2-3} \cmidrule(lr){4-5}
 & $BM$ & $RPD_{BM}$ & $BM$ & $RPD_{BM}$ \\
 \midrule
SLABC (2022)	&	192.8 	&	33.90\% & - & - \\
SLGA (2020)	&	181.3 	&	22.93\% & - & - \\
HGIN-RS (2024)	&	182.5 	&	27.57\% & - & - \\
GRU-DRL (2025)	&	179.5 	&	23.61\%  & - & - \\
HTS/SA (2007) & - & -	&	518.70 	&	19.92\% \\
AIA (2010) & - & -	&	490.20 	&	14.54\% \\
IGAR (2021) & - & -	&	490.45 	&	14.49\% \\
DDEA-PMI (2025) & - & -	&	554.15 	&	25.24\% \\
GA	&	176.2	&	19.13\% 	&	487.95 	&	14.09\% 	\\
LLM4EO	&	\textbf{175.7} 	&	\textbf{18.54\%} 	&	\textbf{487.75} 	&	\textbf{14.06\%} 	\\
\bottomrule
\end{tabular}
\label{table7}
\end{table}

\section{Conclusion}
We propose LLM4EO that utilizes LLM-driven online operator design to enhance the optimization performance of Evolutionary Algorithms (EAs). By leveraging prior knowledge of problem structures and heuristics, it construct high-quality gene selection strategies of operators. When population evolution stagnates, the LLM4EO perceives evolutionary states, analyzes operator limitations, and provides improvement suggestions. The resulting new operators are tailored to the current search stage and guide the EAs to explore potential neighborhoods. On different FJSP benchmarks, LLM4EO outperforms other mainstream algorithms, demonstrating its superior performance. LLM-driven meta-operators consistently accelerate population evolution and enhance solution quality, even under limited iterations.

\appendix



\bibliographystyle{named}
\bibliography{ijcai26}

\appendix
\onecolumn

\section{Algorithm Details}
In this part, we elaborate on the main framework, gene selection and critical operation swapping used in LLM4EO, as illustrated in Algorithm \ref{algorithm3}, Algorithms \ref{algorithm4} and Algorithms \ref{algorithm5}, respectively.

\paragraph{Main Framework.} The solution-operator co-evolution framework of LLM4EO, as presented in Algorithm \ref{algorithm3}, employs GA to optimize the solution population and leverages LLMs to autonomously improve operators in iterations to accelerate optimization. First, the initial solution and operator populations are generated during the initialization phase. Then, new solution population is produced through crossover, mutation, and critical operation swapping. In this process, new individuals are generated by operators that select which genes to perturb. When a predefined threshold is reached, the operator evolution mechanism is triggered, leveraging the LLM to design new operators and replace poor performers within the operator population. Finally, output the optimal solution once the termination condition is satisfied.

\begin{algorithm*}[h]
\caption{The main framework of LLM4EO}
\label{algorithm3}
\textbf{Input}: solution population size $p_{soln}$, operator population size $p_{op}$ \\
\textbf{Output}: the best solution $s'$
\begin{algorithmic}[1] 
\STATE Initialize a solution population $P_{soln}=\{s_1,s_2,...s_{p_{soln}}\}$ and an operator population $P_{op}=\{o_1,o_2,...o_{p_{op}}\}$ \\
\WHILE {the termination condition is satisfied}
\STATE $P'_{soln} \gets \varnothing$
\STATE $s' \gets$ select the best solution among $P_{soln}$
\STATE // Crossover
\FOR {$i=\{1,2,...,p_{soln}/2\}$}
\STATE Select two solutions $s_{x}$ and $s_{y}$ from $P_{soln}$
\STATE Randomly generate a decimal $r_c\in[0,1]$
\STATE // $p_c$ is the crossover probability
\IF {$r_c < p_c$}
\STATE Randomly generate a 0-1 variable $v_c$
\IF{$v_c = 0$}
\STATE Select two gene sets $\delta_x$ and $\delta_y$ of $s_{x}$ and $s_{y}$ based on job features
\STATE $P'_{soln} = P'_{soln} \cup$ Crossover\_OSV($s_{x}, s_{y}, \delta_x,\delta_y$)
\ELSE
\STATE Select two gene sets $\delta_x$ and $\delta_y$ of $s_{x}$ and $s_{y}$ based on operation features
\STATE $P'_{soln} = P'_{soln} \cup$ Crossover\_MAV($s_{x}, s_{y}, \delta_x,\delta_y$)
\ENDIF
\ELSE
\STATE $P'_{soln} = P'_{soln} \cup (s_{x}, s_{y})$
\ENDIF
\ENDFOR
\STATE // Mutation
\FOR {$i=\{1,2,...,p_{soln}\}$}
\STATE Randomly generate a decimal $r_m\in[0,1]$
\STATE // $p_m$ is the mutation probability
\IF {$r_m < p_m$}
\STATE Randomly generate a 0-1 variable $v_m$
\IF{$v_m = 0$}
\STATE Select a gene set $\delta'_i$ of $s'_{i}$ based on job features \quad // $P'_{soln} = \{s'_{1},s'_{2},...,s'_{p_{soln}} \}$
\STATE $s'_{i} \gets $ Mutation\_OSV($s'_{i},\delta'_i$)
\ELSE
\STATE Select a gene set $\delta'_i$ of $s'_{i}$ based on operation features
\STATE $s'_{i} \gets $ Mutation\_MAV($s'_{i},\delta'_i$)
\ENDIF
\ENDIF
\ENDFOR
\STATE // Critical operation swapping
\FOR {$i=\{1,2,...,p_{soln}\}$}
\STATE $s'_{i} \gets $ CriticalOperationSwapping($s'_{i}$, $P_{op}$)
\ENDFOR
\STATE // Population update
\STATE $s_b \gets$ select the best solution among $P_{soln}$
\STATE $s'_w \gets$ select the worst solution among $P'_{soln}$
\STATE $s'_w \gets s_b, P_{soln} \gets P'_{soln}$
\IF {$C_{max}(s_b) < C_{max}(s')$}
\STATE $s' \gets s_b$
\ENDIF
\STATE // Operator evolution by LLMs
\IF {the operator evolution condition is satisfied}
\STATE Generate a new operator $o'$ using the operator evolution procedure.
\STATE $o_w \gets$ select the worst operator among $P_{op}$
\STATE $o_w \gets o'$
\ENDIF
\ENDWHILE
\STATE \textbf{return} $ s' $
\end{algorithmic}
\end{algorithm*}

\paragraph{Gene Selection.} For gene selection, as shown in Algorithm \ref{algorithm4}, features of jobs and operations are leveraged by LLM-driven operators to create linear or non-linear relationships between genes, generating probabilistic distributions. The potential genes are more likely to be selected for perturbation, thereby generating more diverse and superior solutions.

\begin{algorithm*}[h]
\caption{Gene selection}
\label{algorithm4}
\textbf{Input}: solution $s$, operator population $P_{op}$, feature type \{job, operation\} \\
\textbf{Output}: selected genes   $\delta$
\begin{algorithmic}[1] 
\STATE // $I$ represent the numbers of jobs and $J_i$ is the operation count of job $i$, respectively
\STATE // $n$ and $m$ denote the dimensional numbers of job features and operation features, respectively
\STATE Compute job features $\mathcal{N}_{job}=[\{\beta^{1}_1,\beta^{1}_2,...,\beta^{1}_I\},...,\{\beta^{n}_1,\beta^{n}_2,...,\beta^{n}_I\}]$ of solution $s$ 
\STATE Compute operation features $\mathcal{N}_{operation}=[\{\beta^{1}_{1,1},\beta^{1}_{1,2},...,\beta^{1}_{I,J_I}\},...,\{\beta^{m}_{1,1},\beta^{m}_{1,2},...,\beta^{m}_{I,J_I}\}]$ of solution $s$
\STATE Select an operator $o$ from $P_{op}$ by the roulette wheel method
\STATE // CalculatePriority is the function of operator $o$
\STATE $\{\alpha_1,\alpha_2,...,\alpha_I\},\{\alpha_{1,1},\alpha_{1,2},...,\alpha_{I,J_I}\} \gets$ CalculatePriority($\mathcal{N}_{job}, \mathcal{N}_{operation}$)
\STATE $\delta \gets \varnothing$
\IF {feature type is job}
\STATE $\{\gamma_1,\gamma_2,...\gamma_I\} \gets$ Normalisation($\{\alpha_1,\alpha_2,...\alpha_I\}$)
\FOR {$i \leq I$}
\STATE Randomly generate a decimal $u_i\in[0,1]$
\IF {$\gamma_i > u_i$}
\STATE $\delta' \gets$ select all genes of operations belonging to the job $i$
\STATE $\delta  = \delta  \cup \delta'$
\ENDIF
\ENDFOR
\ELSIF{feature type is operation}
\STATE $\{\gamma_{1,1},\gamma_{1,2},...\gamma_{I,J_I}\} \gets$ Normalisation($\{\alpha_{1,1},\alpha_{1,2},...,\alpha_{I,J_I}$)
\FOR {$O_{ij}(i \leq I, j \leq J_i))$}
\STATE Randomly generate a decimal $u_{ij}\in[0,1]$
\IF {$\gamma_{ij} > u_{ij}$}
\STATE $\delta' \gets$ select the gene of operation $O_{ij}$
\STATE $\delta  = \delta  \cup \delta'$
\ENDIF
\ENDFOR
\ENDIF
\STATE \textbf{return} $ \delta $
\end{algorithmic}
\end{algorithm*}

\paragraph{Critical Operation Swapping.} For critical operation swapping, as shown in Algorithm \ref{algorithm5}, genes of the critical path are selected by meta-operators and then two of them are randomly exchanged. If a better individual is obtained, the search process is repeated on the new critical path. Otherwise, the process is terminated. This approach combining local search with LLM-driven operators helps accelerate convergence.

\begin{algorithm}[h]
\caption{Critical operation swapping}
\label{algorithm5}
\textbf{Input}: solution $s$, operator population $P_{op}$ \\
\textbf{Output}: the best solution $s'$
\begin{algorithmic}[1] 
\STATE $s' \gets s$
\WHILE {$s'$ achieves the lower makespan}
\STATE $s \gets s'$
\STATE Identify the critical path of $s$
\STATE Select an operator $o$ from $P_{op}$ by roulette
\STATE Select two genes $g_x$ and $g_y$ of the critical path by operator $o$
\STATE Swap $g_x$ and $g_y$
\IF {$C_{max}(s) < C_{max}(s')$}
\STATE $s' \gets s$
\ENDIF
\ENDWHILE
\STATE \textbf{return} $ s' $
\end{algorithmic}
\end{algorithm}

\section{Prompt and Generation Details}
This section provides examples of the carefully crafted prompts and LLM responses for each module of the co-evolutionary architecture. These modules include operator design, evolutionary perception and analysis, and adaptive operator evolution.

\subsection{Operator Design}
\definecolor{color1}{RGB}{112,91,56}
\definecolor{color2}{RGB}{252,119,39}
\definecolor{color3}{RGB}{0,102,204}
\definecolor{color4}{RGB}{112,48,160}
\definecolor{color5}{RGB}{192,0,0}

In this subsection, we introduce the prompt for operator initialization and the details of the initial operators, as shown in Figure \ref{Appendix_Operator_Initialization}. The five different colors represent five different components of the prompt, including \textcolor{color1}{Task description}, \textcolor{color2}{Prior knowledge}, \textcolor{color3}{Expected output}, \textcolor{color4}{Code template}, and \textcolor{color5}{Special hints}. Based on the structured prompt, the LLM generates three distinct initial operators and provides descriptions of these, as well as executable functions that follow the code template. These operators establish linear or nonlinear relationships among genes based on seven genetic features and output gene selection priorities.

\subsection{Evolution Perception and Analysis}

\definecolor{color6}{RGB}{214,138,24}
\definecolor{color7}{RGB}{0,138,63}

This subsection presents the prompt and the result for the perception and analysis of evolution in Figure \ref{Appendix_Perception_and_Analysis}. The five different colors represent five different components of the prompt, including \textcolor{color6}{Solution population changes}, \textcolor{color7}{Operator population information}, \textcolor{color1}{Task description}, \textcolor{color3}{Expected output}, and \textcolor{color5}{Special hints}. Population changes can be described using metrics such as the iteration count, minimum value, average value, and rate of change. For the operator population performance, the fitness value and the computational approach of each operator are described. Given the above information, the LLM is expected to assess the optimization dilemma, analyze the limitations of the operator population, and provide a design suggestion.

\subsection{Adaptive Operator Evolution}
As shown in Figure \ref{Appendix_Operator_Evolution}, most of the content in the prompts for adaptive operator evolution comes from operator design, as well as evolutionary perception and analysis. Notably, the task descriptions are derived from variations of the initial operator design task via the LLM, which helps mitigate the inertia of the LLM and generate diverse operators, as illustrated in Figure \ref{Appendix_Task_Mutation}. According to the suggestions provided by the LLM, the priority calculation rules for new operators focus on process order, start time, and processing time, as indicated by the bolded content in the code. It emphasizes \enquote{sequential bottlenecks and machine substitution potential to minimize makespan}.

\section{Other Experimental Details}

\subsection{Comparison with Genetic Algorithms}
To further evaluate the impact of LLM-generated meta-operators on the convergence of evolutionary algorithms, LLM4EO is compared with several genetic algorithms. These algorithms include the traditional genetic algorithm proposed by F.Pezzella et al. (P'GA) \cite{r:8}, the self-learning genetic algorithm based on reinforcement learning (SLGA)  \cite{r:40}, and the basic genetic algorithm (GA) of LLM4EO. The components of GA and LLM4EO are essentially the same, with the only difference being that GA generates offspring by randomly selecting genes rather than relying on LLM-generated operators. The operator population size is set to 3, the solution population size to 100 and the maximum number of iterations to 200. These algorithms run independently 10 times on each Brandimarte benchmark instance. 

As shown in Table \ref{tableS1}, LLM4EO achieves optimal $BM$ and $AM$ values for all instances except MK07, which demonstrates its outstanding performance. Furthermore, the average relative percentage deviation ($RPD_{aver}$) of LLM4EO is lower than that of the comparison algorithms, thus validating its effectiveness in solving the FJSP. Notably, LLM4EO is based on GA but employs meta-operators to select promising genes for generating individuals. With the assistance of LLMs, the $RPD_{aver}$ of $BM$ decreased from 13.19 to 12.71, achieving a 3.64\% performance improvement; the $RPD_{aver}$ of $AM$ dropped from 14.58 to 14.12, yielding a 3.16\% performance gain. These improvements demonstrate that the proposed meta-operator effectively enhances the exploration capability and stability of evolutionary algorithms.

\begin{figure*}[h]
\centering
\includegraphics[width=1\columnwidth]{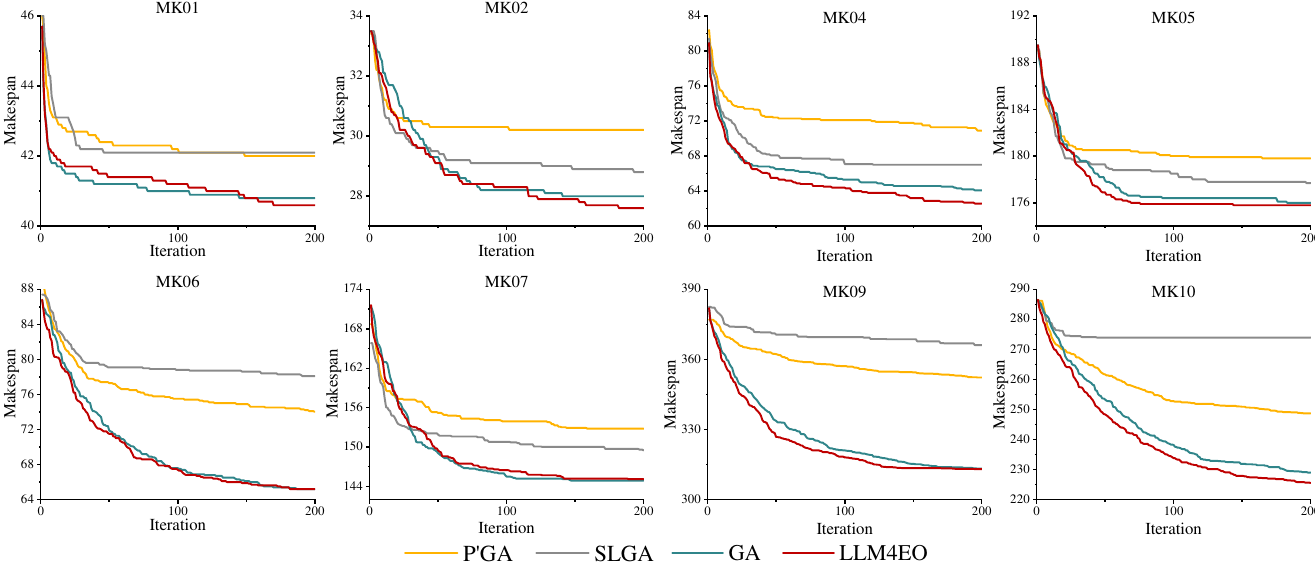} 
\caption{Convergence curves of genetic algorithms on Brandimarte benchmark.}
\label{Fig_GA_convergence}
\end{figure*}

Figure \ref{Fig_GA_convergence} presents the average convergence of four algorithms after running 10 times on the MK01, MK02, MK04, MK05, MK06, MK07, MK09, and MK10 instances. For the MK03 and MK08 instances, no significant convergence process is observed since the initial solutions already attain the theoretical lower bound. It can be seen that, in most cases, LLM4EO converges faster than other algorithms and ultimately yields higher-quality solutions. This phenomenon validates that meta-operators based on LLMs effectively accelerate the algorithmic evolution process.

\begin{table*}[h]
\caption{Comparison results of genetic algorithms with LLM4EO.}
\centering
\setlength{\tabcolsep}{1mm}
\fontsize{9pt}{9pt}\selectfont
\begin{tabular}{ccccccccccc}
\toprule
\multirow{2}{*}{Instance} & \multirow{2}{*}{$Size$} & \multirow{2}{*}{$LB$} & \multicolumn{4}{c}{$BM$} & \multicolumn{4}{c}{$AM$} \\
\cmidrule(lr){4-7} \cmidrule(lr){8-11}
 & & &	F'GA	&	SLGA	&	GA	&	LLM4EO	&	F'GA	&	SLGA	&	GA	&	LLM4EO \\
 \midrule
MK01	&	10×6	&	36 	&	42 	&	42 	&	\textbf{40} 	&	\textbf{40} 	&	42.6 	&	42.1 	&	40.8 	&	\textbf{40.6} 	\\
MK02	&	10×6	&	24 	&	29 	&	28 	&	\textbf{27} 	&	\textbf{27} 	&	30.6 	&	28.8 	&	28.0 	&	\textbf{27.6} 	\\
MK03	&	15×8	&	204 	&	204 	&	204 	&	204 	&	204 	&	204.0 	&	204.0 	&	204.0 	&	204.0 	\\
MK04	&	15×8	&	48 	&	69 	&	66 	&	62 	&	\textbf{60} 	&	72.0 	&	67.0 	&	64.1 	&	\textbf{62.6} 	\\
MK05	&	15×4	&	168 	&	177 	&	176 	&	174 	&	\textbf{173} 	&	178.5 	&	177.7 	&	176.0 	&	\textbf{175.8} 	\\
MK06	&	10×15	&	33 	&	71 	&	74 	&	\textbf{63} 	&	\textbf{63} 	&	75.1 	&	78.1 	&	\textbf{65.2} 	&	\textbf{65.2} 	\\
MK07	&	20×5	&	133 	&	150 	&	145 	&	\textbf{143} 	&	144 	&	154.1 	&	149.5 	&	\textbf{144.9} 	&	145.1 	\\
MK08	&	20×10	&	523 	&	523 	&	523 	&	523 	&	523 	&	527.4 	&	\textbf{523.0} 	&	\textbf{523.0} 	&	\textbf{523.0} 	\\
MK09	&	20×10	&	299 	&	362 	&	360 	&	\textbf{311} 	&	\textbf{311} 	&	368.1 	&	366.1 	&	313.2 	&	\textbf{313.0} 	\\
MK10	&	20×15	&	165 	&	251 	&	269 	&	224 	&	\textbf{217} 	&	258.7 	&	273.9 	&	229.0 	&	\textbf{225.5} 	\\
\midrule
$RPD_{aver}$ & & &	18.36 	&	17.97 	&	13.19 	&	\textbf{12.71} 	&	20.19 	&	19.18 	&	14.58 	&	\textbf{14.12} \\
\bottomrule
\end{tabular}
\label{tableS1}
\end{table*}

\subsection{LLM4EO on Distributed Flexible Job Shop Scheduling}
The experimental scope is expanded to the distributed flexible job shop scheduling problem (DFJSP), with the aim of exploring the generalization and robustness of LLM4EO. The benchmark from \cite{r:45} is used to test performance. Three comparison algorithms include Giovanni'GA  \cite{r:45}, Lu'GA \cite{r:46}, CRO \cite{r:52} and the basic genetic algorithm (GA) of LLM4EO. 
Three key components of the proposed algorithm are adjusted:
1) Add factory crossover: Select multiple operations from the parent and swap their factories to generate new offspring. 
2) Add factory mutation: Assign the selected operations to other machines of the current factory. 
3) Modified critical operation swapping: In the critical path, select an operation and sequentially swap it with other operations. If fitness improves, repeat this local search process on the new critical path.

\begin{figure*}[h]
\centering
\includegraphics[width=0.7\columnwidth]{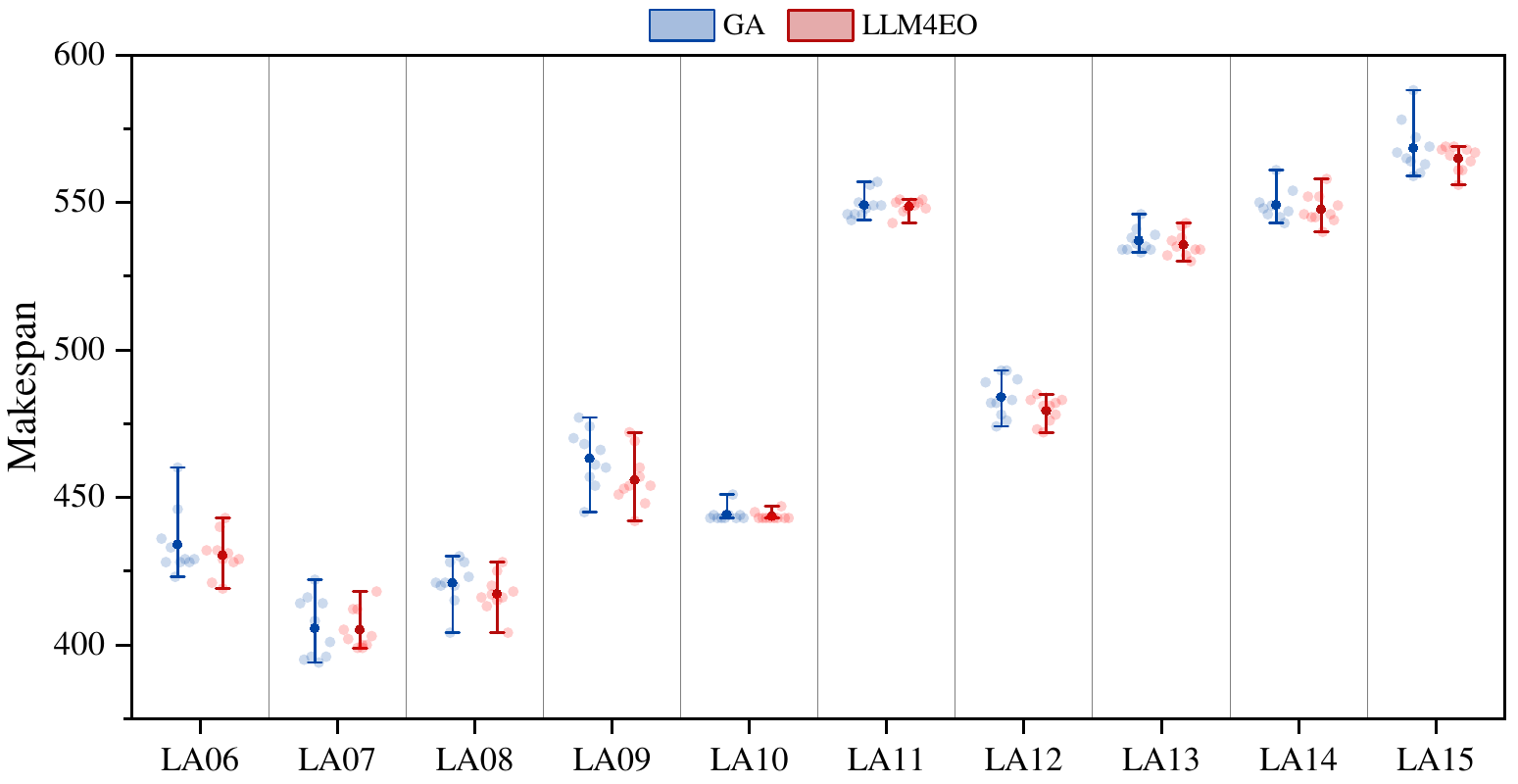} 
\caption{Partial box plots of LLM4EO and GA on LA06-LA15 instances.}
\label{Fig_box_plots_for_DFJSP}
\end{figure*}

\begin{figure*}[h]
\centering
\includegraphics[width=1\textwidth]{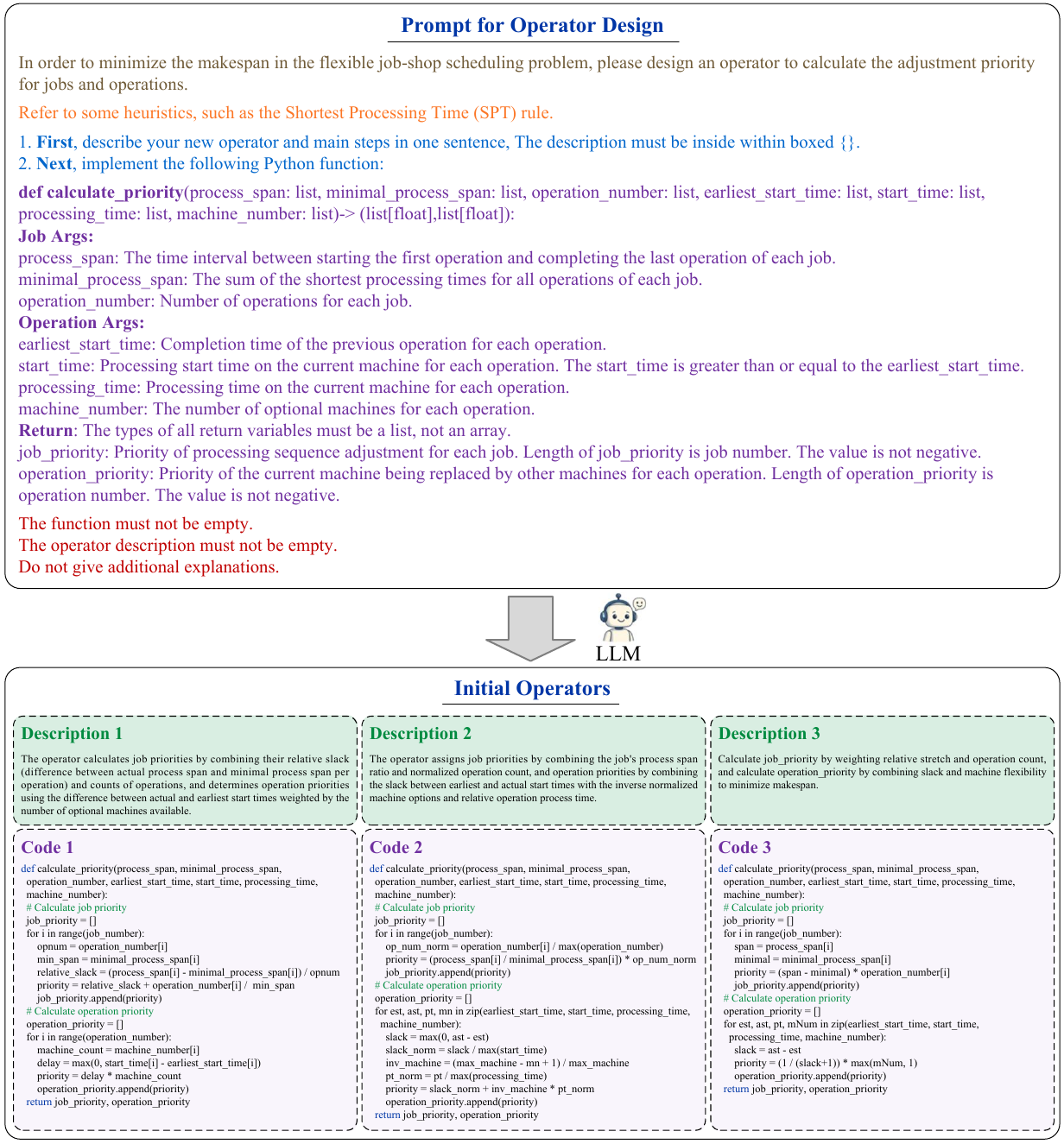}
\caption{Prompt and output for initial operator design.}
\label{Appendix_Operator_Initialization}
\end{figure*}

\begin{figure*}[h]
\centering
\includegraphics[width=1\textwidth]{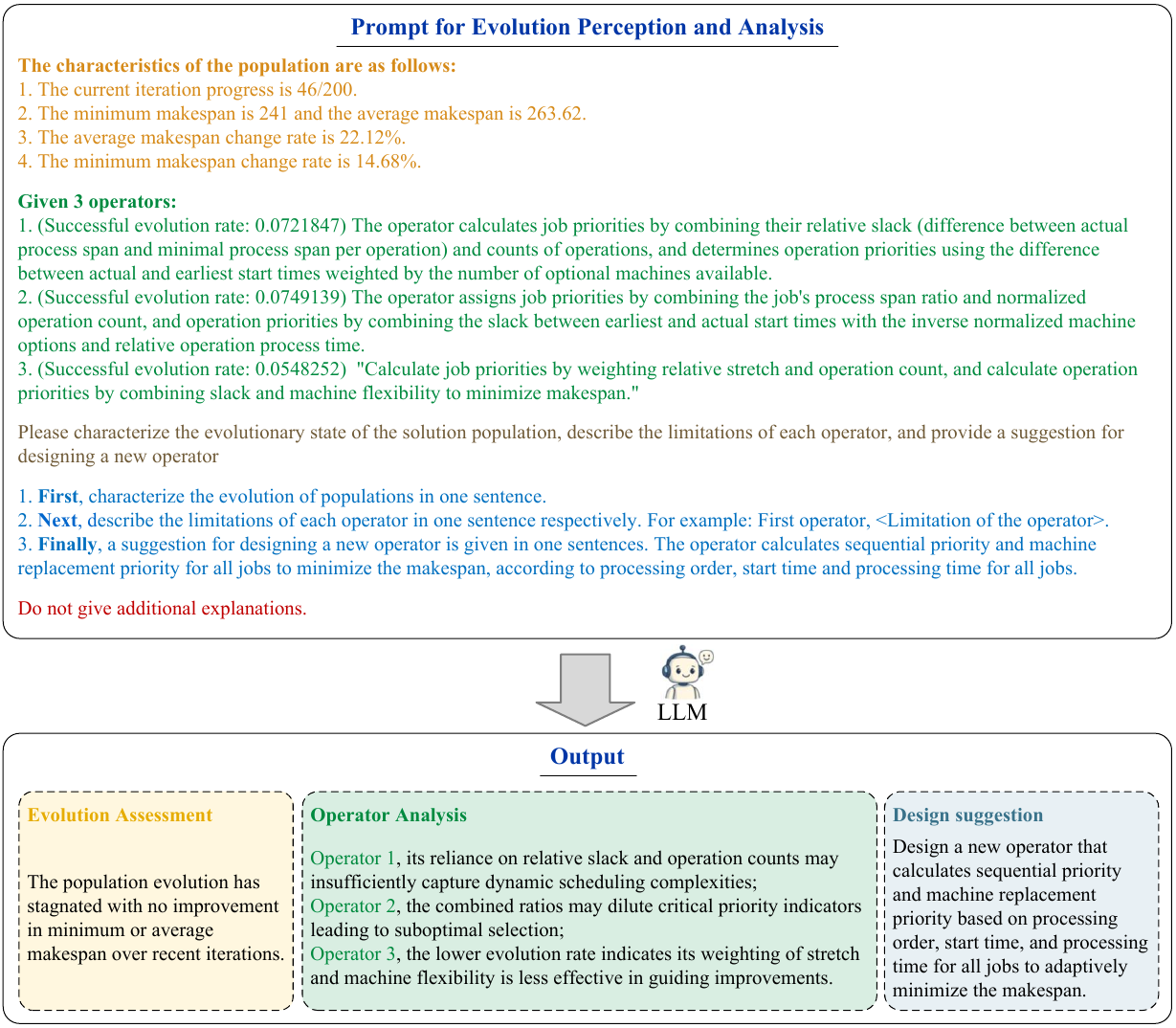}
\caption{Prompt and output for evolution perception and analysis.}
\label{Appendix_Perception_and_Analysis}
\end{figure*}

\begin{figure*}[h]
\centering
\includegraphics[width=1\textwidth]{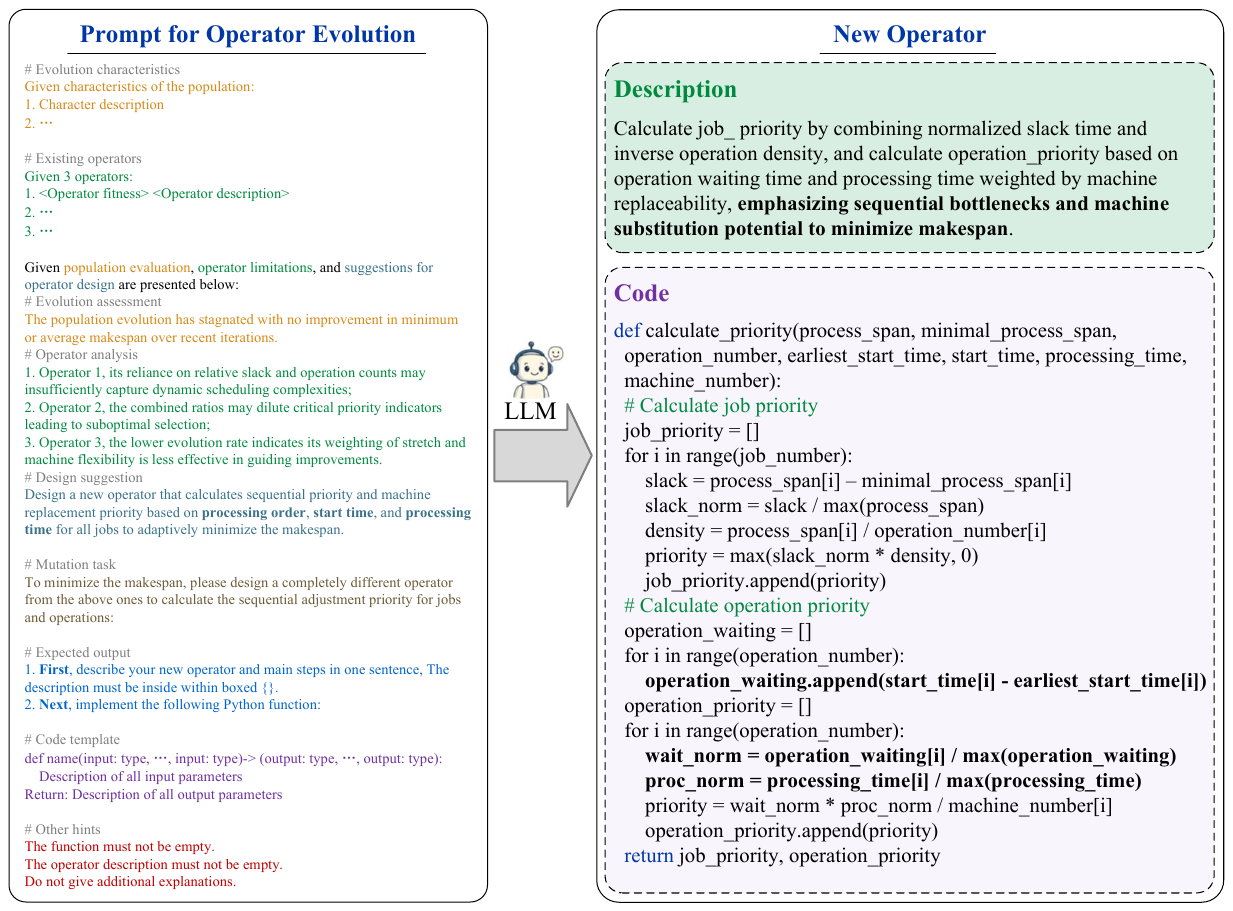}
\caption{Prompt and output for operator evolution.}
\label{Appendix_Operator_Evolution}
\end{figure*}

\begin{figure}[h]
\centering
\includegraphics[width=0.7\columnwidth]{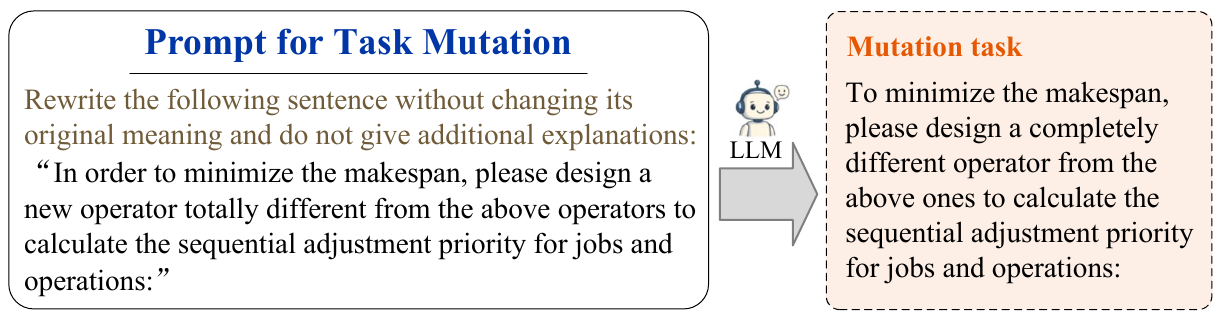}
\caption{Prompt and output for task mutation.}
\label{Appendix_Task_Mutation}
\end{figure}

Table \ref{tableS2} presents the results for 20 instances with two factories. It can be seen that all algorithms achieve $LB$ in LA01-LA05 and LA16-LA20. Additionally, LLM4EO obtains the optimal $BM$ in LA06 and LA8-LA15, demonstrating its outstanding generalization. Figure~\ref{Fig_box_plots_for_DFJSP} shows the box plot of the makespan values obtained by LLM4EO and GA in ten runs. It is evident that the result distribution of LLM4EO is more concentrated than that of GA, with a shorter box positioned closer to the lower end. This indicates that it exhibits superior performance and fewer fluctuations in most cases. Therefore, the proposed LLM-driven meta-operator is still effective in enhancing the performance of evolutionary algorithms when solving the DFSP.

\begin{table*}[h]
\caption{Computational results for the DFJSP with two factories.}
\centering
\setlength{\tabcolsep}{1mm}
\fontsize{9pt}{9pt}\selectfont
\begin{tabular}{cccccccccccc}
\toprule
\multirow{2}{*}{Instance} & \multirow{2}{*}{$Size$} & \multirow{2}{*}{$LB$}  & \multicolumn{2}{c}{LLM4EO} & \multicolumn{2}{c}{GA} & \multicolumn{2}{c}{Giovanni'GA} & \multicolumn{2}{c}{Lu'GA} & \multirow{2}{*}{CRO} \\
\cmidrule(lr){4-5} \cmidrule(lr){6-7} \cmidrule(lr){8-9} \cmidrule(lr){10-11} 
 &   &  & $BM$ & $AM$ & $BM$ & $AM$ & $BM$ & $AM$ & $BM$ & $AM$ & \\
 \midrule
LA01 &  10 × 5 &	413	&	413	&	413	&	413	&	413	&	413	&	413	&	413	&	413	&	413	\\
LA02 &  10 × 5 &	394	&	394	&	394	&	394	&	394	&	394	&	394	&	394	&	394	&	394	\\
LA03 &  10 × 5 &	349	&	349	&	349	&	349	&	349	&	349	&	349	&	349	&	349	&	349	\\
LA04 &  10 × 5 &	369	&	369	&	369	&	369	&	369	&	369	&	369	&	369	&	369	&	369	\\
LA05 &  10 × 5 &	380	&	380	&	380	&	380	&	380	&	380	&	380	&	380	&	380	&	380	\\
LA06 &  15 × 5 &	413	&	\textbf{419}	&	430.4	&	423	&	434	&	445	&	449.6	&	424	&	435.8	&	424	\\
LA07 &  15 × 5 &	376	&	399	&	405	&	\textbf{394}	&	405.6	&	412	&	419.2	&	398	&	408.5	&	398	\\
LA08 &  15 × 5 &	369	&	\textbf{404}	&	417.2	&	\textbf{404}	&	421	&	420	&	427.8	&	406	&	417.4	&	406	\\
LA09 &  15 × 5 &	382	&	\textbf{442}	&	456	&	445	&	463.2	&	469	&	474.6	&	447	&	459	&	463	\\
LA10 &  15 × 5 &	443	&	\textbf{443}	&	443.6	&	\textbf{443}	&	444.1	&	445	&	448.6	&	\textbf{443}	&	444.1	&	445	\\
LA11 &  20 × 5 &	413	&	\textbf{543}	&	548.7	&	544	&	549.1	&	570	&	571.6	&	548	&	557.1	&	553	\\
LA12 &  20 × 5 &	408	&	\textbf{472}	&	479.4	&	474	&	484	&	504	&	508	&	480	&	492.5	&	500	\\
LA13 &  20 × 5 &	382	&	\textbf{530}	&	535.7	&	533	&	537	&	542	&	552.2	&	533	&	538.4	&	551	\\
LA14 &  20 × 5 &	443	&	\textbf{540}	&	547.7	&	543	&	549.2	&	570	&	576	&	542	&	557.3	&	581	\\
LA15 &  20 × 5 &	378	&	\textbf{556}	&	564.9	&	559	&	568.5	&	584	&	588.8	&	562	&	568.7	&	597	\\
LA16 &  10 × 10 &	717	&	717	&	717	&	717	&	717	&	717	&	717	&	717	&	717	&	717	\\
LA17 &  10 × 10 &	646	&	646	&	646	&	646	&	646	&	646	&	646	&	646	&	646	&	646	\\
LA18 &  10 × 10 &	663	&	663	&	663	&	663	&	663	&	663	&	663	&	663	&	663	&	663	\\
LA19 &  10 × 10 &	617	&	617	&	617	&	617	&	617	&	617	&	617.2	&	617	&	622.1	&	617	\\
LA20 &  10 × 10 &	756	&	756	&	756	&	756	&	756	&	756	&	756	&	756	&	756	&	756	\\
\bottomrule
\end{tabular}
\label{tableS2}
\end{table*}

\clearpage
\newpage



\end{document}